\documentclass[letterpaper]{article} 
\usepackage{aaai25}  
\usepackage{times}  
\usepackage{helvet}  
\usepackage{courier}  
\usepackage[hyphens]{url}  
\usepackage{graphicx} 
\usepackage{natbib}  
\usepackage{caption} 
\usepackage{amsmath}

\usepackage{algorithm}
\usepackage{algorithmic}
\usepackage{rotating}
\usepackage{makecell}
\usepackage{xcolor}
\usepackage{amssymb}

\usepackage{newfloat}
\usepackage{listings}
\DeclareCaptionStyle{ruled}{labelfont=normalfont,labelsep=colon,strut=off} 
\floatstyle{ruled}
\newfloat{listing}{tb}{lst}{}
\floatname{listing}{Listing}
\nocopyright 

\title{GINN-KAN: Interpretability pipelining with applications in Physics Informed Neural Networks}
\author {
    Nisal Ranasinghe\textsuperscript{\rm 1},
    Yu Xia\textsuperscript{\rm 1},
    Sachith Seneviratne\textsuperscript{\rm 1},
    Saman Halgamuge\textsuperscript{\rm 1}
}
\affiliations {
    \textsuperscript{\rm 1} AI, Optimization and Pattern Recognition Research Group, Dept. of Mechanical Eng., University of Melbourne, Australia\\
    nsranasinghe@student.unimelb.edu.au, yu.xia8@unimelb.edu.au, sachith.seneviratne@unimelb.edu.au, saman@unimelb.edu.au
}

\begin{document}

\maketitle

\begin{abstract}
Neural networks are powerful function approximators, yet their ``black-box" nature often renders them opaque and difficult to interpret. While many post-hoc explanation methods exist, they typically fail to capture the underlying reasoning processes of the networks. A truly interpretable neural network would be trained similarly to conventional models using techniques such as backpropagation, but additionally provide insights into the learned input-output relationships. In this work, we introduce the concept of interpretability pipelineing, to incorporate multiple interpretability techniques to outperform each individual technique. To this end, we first evaluate several architectures that promise such interpretability, with a particular focus on two recent models selected for their potential to incorporate interpretability into standard neural network architectures while still leveraging backpropagation: the Growing Interpretable Neural Network (GINN) and Kolmogorov Arnold Networks (KAN). We analyze the limitations and strengths of each and introduce a novel interpretable neural network GINN-KAN that synthesizes the advantages of both models. When tested on the Feynman symbolic regression benchmark datasets, GINN-KAN outperforms both GINN and KAN. To highlight the capabilities and the generalizability of this approach, we position GINN-KAN as an alternative to conventional black-box networks in Physics-Informed Neural Networks (PINNs), which we propose as a challenging testbed for backpropagation-friendly interpretable neural networks. By adding interpretability to PINNs, we allow far more transparent and trustworthy data-driven solutions to differential equations. We expect this to have far-reaching implications in the application of deep learning pipelines in the natural sciences. Our experiments with this interpretable PINN on 15 different partial differential equations demonstrate that GINN-KAN augmented PINNs outperform PINNs with black-box networks in solving differential equations and surpass the capabilities of both GINN and KAN.
\end{abstract}

%

\section{Introduction}


Neural networks have largely driven the advancement of artificial intelligence (AI) over the last decade. They are widely used in many domains including computer vision, natural language processing and speech processing for real-world applications. Such neural networks have already surpassed human performance in many applications. However, neural networks generally consist of a large number of neurons, which results in them learning a complex function that maps the inputs to the outputs. Though this function may be very accurate, their innate opaque nature prevents it from being adopted in many sensitive fields where critical decisions have to be made using the output of the model, and the ability to describe and justify the decision becomes paramount. 

The interpretability of neural networks is crucial to increasing the trustworthiness of AI, thereby increasing its adoption in real-world applications. Though explainability of machine learning (ML) models including neural networks have been studied in the past \cite{ribeiro_why_2016, lundberg_unified_2017, zhou_learning_2016, selvaraju_grad-cam_2016}, there is a subtle difference between explainability and interpretability. Explainability focuses on explaining the decisions made by a system and typically does not represent how the model actually made its decision \cite{ali_explainable_2023}. On the other hand, an interpretable model aims to provide insights into the reasoning of the model.

In this work, we focus on interpretable neural networks, which give insights into the learned representation of a model. In particular, we are interested in networks that can aid in scientific discovery, by learning functions which follow a concise mathematical equation while still being trained on input-output pairs similar to any other neural network. The growing interpretable neural network (GINN) \cite{ranasinghe_ginn-lp_2024} and the Kolmogorov Arnold Network (KAN) \cite{liu_kan_2024} are two such networks that have shown promise in this task, and have even been shown to be able to discover the ground truth mathematical equations while still being trained using regular backpropagation. The ability to learn equations is similar to symbolic regression (SR) but with a key difference. A SR algorithm searches the equation space in an efficient manner until the best equation describing the dataset is found. In contrast, these methods can be trained on input-output pairs like any other neural network, but in a way that allows the extraction of a mathematical equation upon inspection of the trained neural network. 

Leveraging this unique property exhibited by GINN and KANs, we introduce the concept of interpretability pipelining, which combines multiple interpretability methods to enhance model transparency and performance. We present GINN-KAN, the first such pipelined interpretable neural network, that combines the strengths of the GINN and KAN architectures, resulting in a far more robust network, while retaining interpretability.

A promising application for such interpretable neural networks lies in the field of Physics Informed Neural Networks (PINN) \cite{raissi_physics_2017}. PINNs make use of regularization and auto-differentiation to enforce a neural network to fit the underlying partial differential equation. Although PINNs have shown great potential in solving partial differential equations (PDEs), their reliance on black-box neural networks results in black-box solutions to these PDEs, significantly limiting the understanding of these solutions. By incorporating GINN-KAN into the PINN framework, we propose a new type of interpretable PINN which can provide valuable insights into the analytical solution of a PDE. Although this has been briefly investigated in \cite{liu_kan_2024}, a comprehensive evaluation of such interpretable PINNs has not been performed.

In this work, we investigate the strengths and weaknesses of current backpropagation-friendly interpretable neural networks. We then use these insights to propose a new type of interpretable network that combines the strengths of these methods and use this to create an interpretable physics-informed neural network. The \textbf{main contributions} of this work are,
\begin{itemize}
    \item An evaluation of interpretable neural networks, including KANs on symbolic regression datasets
    \item GINN-KAN: A new type of interpretable neural network that can aid in scientific discovery
    \item Integration of interpretability into PINNs allowing for better inspection/extraction/interoperability of PINN solutions
    \item An interpretable PINN using GINN-KAN and its evaluation across multiple different PDEs
\end{itemize}

\section{Neural Network Interpretability}

Machine learning model explainability has garnered considerable academic interest. This has resulted in many explainability methods that can visually or quantitatively explain a model's output. Although these methods provide insights into the decisions of the model, they may not reflect the underlying decision-making process. Therefore, even with incorporated explainability methods, machine learning models, especially neural networks remain as black-boxes. On the other hand, a fully interpretable model can be considered a ``white-box", since its inner workings are fully transparent and human-understandable. Some simple models like linear regressors and decision trees are inherently interpretable since the inner workings of the learned model can be simply interpreted using a mathematical equation (in linear regression) or a set of rules (in decision trees). However, these simple models cannot fit datasets with complex underlying functions. 

Neural networks such as GINN \cite{ranasinghe_ginn-lp_2024}, KAN \cite{liu_kan_2024} and EQL \cite{sahoo_learning_2018} can be considered interpretable since they provide insights into the learned function, once trained. They have also been shown to be able to discover mathematical equations that describe the inner workings of the model while being trained using backpropagation. 

\textbf{Growing interpretable network (GINN)}. GINN is an interpretable neural network that uses logarithmic and exponential activations to discover a multivariate Laurent polynomial (LP) equation mapping the input to the output. The network is fast to train and has only a few parameters, and has been shown to perform well in equation discovery when compared to other symbolic regression methods. Though GINN was created to discover equations with a Laurent polynomial structure, it is still reasonably successful in approximating non-LP equations.

\textbf{Kolmogorov Arnold Networks (KAN)}. The KAN \cite{liu_kan_2024} is a new class of neural network architecture that has emerged as a potential replacement for the widely used multi-layer perceptrons (MLP). Unlike traditional MLPs, which use fixed activation functions on nodes, KANs have learnable activation functions on edges, replacing linear weights with univariate B-spline functions. Other types of functions like wavelet functions have also been shown to be effective with this architecture \cite{bozorgasl_wav-kan_2024}. This design enhances interpretability, allowing KANs to be intuitively visualized and easily interacted with, making them potential tools for scientific discovery. While KANs have been extended for applications such as time series forecasting \cite{vaca-rubio_kolmogorov-arnold_2024} and recommendation models \cite{xu_fourierkan-gcf_2024}, their capability on symbolic regression datasets has not been comprehensively evaluated.

In this work, we focus on GINN and KANs since they have been shown to perform well in learning interpretable functions in the recent literature. Both of them can be trained using backpropagation and output a mathematical equation that describes the learned function. 

\subsection{Limitations of GINN and KAN}
KANs are built around the Kolmogorov-Arnold representation theorem, which states that any multivariate function can be represented as the summation of univariate functions. This allows KANs to learn univariate functions as activations on the edges of the network, which are then summed up at the nodes. KANs can perform equation discovery by replacing each learned activation with a known symbolic function (e.g., $\cos$, $\log$, $x^2$, $x^3$). However, due to the nature of this formulation, KANs need multiple layers to learn a simple multiplication $x_1*x_2$ of the inputs $x_1$, $x_2$. The KAN would need to learn this as $\frac{(x_1 + x_2)^2}{4} + \frac{(x_1 - x_2)^2}{4}$, requiring 2 KAN layers.

In contrast, GINN can very easily learn multiplications using its power-term approximator blocks (PTA) but struggles to learn trigonometric and exponential functions since it does not explicitly include such functions in the network architecture. However, they can reasonably approximate non-LP functions using a polynomial-like approximation, and can also be ensembled with other SR methods to perform better with non-LP equations.

\subsection{Physics-informed Neural Networks}
PINNs are a deep learning method for solving PDEs. They ensure that the outputs comply with known physical laws by integrating domain-specific knowledge as soft constraints into the loss function. This is achieved by including ``physics" loss terms which penalize the neural network from deviating from the underlying governing differential equation. The physics loss is defined as follows:

\begin{equation}
    \label{eq:physics_loss}
    \mathcal{L}_{\text{physics}} = \frac{1}{N} \sum_{i=1}^{N} \left( \mathcal{N}[\hat{u}(x_i,t_i)]\right)^2,
\end{equation}
where $\hat{u}(\cdot)$ is the estimated solution of the PDE and $\mathcal{N}$ represents the linear or nonlinear operator. $x$ and $t$ denote space and time respectively, where $x\in \Omega \subset \mathbb{R}^d,\ t \in [0, T]$.  $T$ is the time horizon, and $\Omega$ is the spatial domain. $N$ is the number of collocation points within the spatiotemporal domain.
This allows the network to be optimized using backpropagation techniques. PINNs generally do not require labelled datasets since the labels used for training the network can be generated using the initial or boundary conditions of the differential equation.

\section{Methodology }

\subsection{GINN-KAN}
\begin{figure*}[h]
    \centering
    \includegraphics[width=\textwidth]{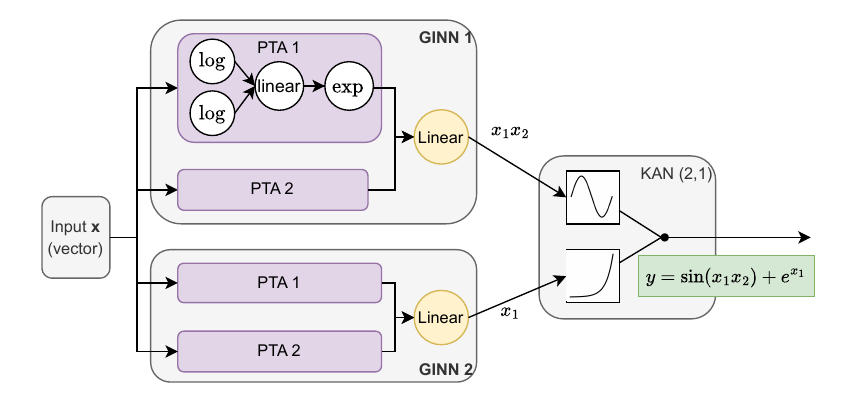}
    \caption{The architecture of GINN-KAN. It consists of two parallel GINN blocks, followed by a KAN block. Each GINN is composed of multiple parallel power-term approximation (PTA) blocks, with each block aiming to discover a single multiplication term in the equation. The entire network is trained end-to-end using backpropagation, which trains the KAN and the PTA blocks within the GINNs. The example shows how GINN-KAN could discover the equation $y = sin(x_1x_2) + e^{x_1} $}
    \label{fig:ginn-kan-arch}
\end{figure*}

We introduce GINN-KAN, an improved interpretable neural network that combines the strengths of GINNs and KANs. While GINNs excel when the underlying ground truth equation is a Laurent polynomial, their performance declines with non-LP equations. Conversely, KANs have the theoretical capability to learn non-LP equations, but in practice struggle with equations involving multiplications due to inherent assumptions. 

We note many parallels between GINNs and KANs, enabling their seamless integration into a more robust network that leverages their strengths while mitigating their individual limitations. Both these networks can be trained using regular backpropagation algorithms on a dataset of input-output pairs. Once trained both GINNs and KANs are interpretable, providing insights into the learned function. Moreover, both of these networks can also use the trained weights of the network to discover a mathematical equation describing the learned function.

We make use of these parallels between GINNs and KANs, to create GINN-KAN, an end-to-end differentiable interpretable neural network, which can be trained using backpropagation. Similar to GINN and KAN, once trained, GINN-KAN can be inspected to gain insights into the network.

The GINN-KAN architecture is shown in Fig. \ref{fig:ginn-kan-arch}. Both the GINN and KAN components of this network can be expressed in terms of a concise mathematical equation. For the GINNs part, this can be done by inspecting the weights and constructing the equation of the network. Each power-term approximator (PTA) block consists of logarithm activations, a single linear activated neuron and an exponential activation. The equation of a PTA block reduces to,
\begin{equation}
    p_i = x_1^{w_{i1}}*x_2^{w_{i2}}*...{x_n^{w_{in}}}
    \label{eq:pta}
\end{equation}
where $p_i$ is the output of the $i$th PTA block. Hence, the equation for a GINN with $n$ power-term approximator (PTA) blocks is given by,

\begin{equation}
   y = \sum_{i=1}^{n} a_i * p_i
   \label{eq:sum_of_powers}
\end{equation}
where $a_i$ is the $i$th coefficient of the linear activated neuron, and $w_{ij}$ are the coefficients of the PTAs.

The KAN can be expressed using an equation by mapping each learned B-spline function to a symbolic function, by comparing them against a set of pre-defined symbolic functions from a library of univariate functions. Some example functions include $\sin(x)$, $e^x$ and $\ln(x)$.

\subsection{GINN-KAN Augmented PINNs}

PINNs are black-box in nature due to the fully connected neural network neural network which is generally used as the function approximator. Though some work attempts to learn symbolic functions that approximate the learned function, this is done by training symbolic regression methods using data generated by black-box surrogate models \cite{podina_universal_2023}. Although this allows the discovery of an interpretable model that approximates the learned function, this may not accurately reflect the decision-making process of the trained PINN.

\begin{figure*}[h]
    \centering
    \includegraphics[width=0.95\textwidth]{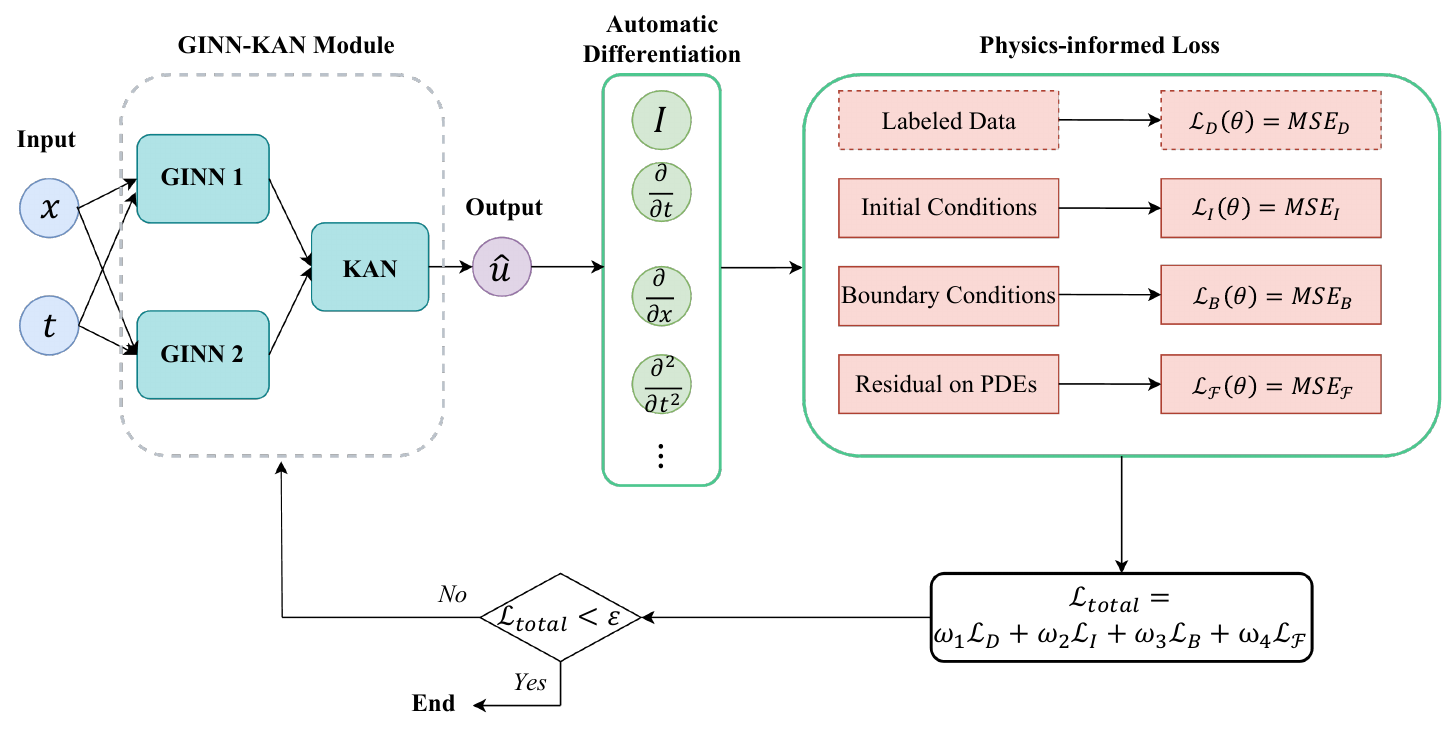}
    \caption{The architecture of GINN-KAN augmented PINNs. The proposed GINN-KAN module replaces the fully connected networks in the conventional PINNs, converting the black-box PINN into an interpretable PINN, without making any changes to the training strategy, since this interpretable PINN can still be trained using backpropagation}
    \label{fig: GINN-KAN with pinn}
\end{figure*}

In this paper, we propose replacing the black-box neural networks within PINNs with GINN-KAN to create an interpretable PINN. Since GINN-KAN can be trained using backpropagation, it can be used within PINNs with minimal change to the training strategy. Once trained, the interpretable PINN will be able to provide insights into the learned solution to the differential equation. The architecture of GINN-KAN augmented PINNs is shown in Fig. \ref{fig: GINN-KAN with pinn}.

\section{Experiment Setup}

\subsection{GINN-KAN}
\textbf{Investigating the limitations of GINN and KANs}. Earlier, we noted that KANs cannot easily approximate multiplications, while GINN cannot easily approximate non-LP equations. To confirm these hypotheses, we construct datasets using eight ground truth equations. Two of these equations are LP equations, two are non-LP equations without any multiplications and the remaining four are non-LP equations with multiplications. We compare the performances of GINN, KAN and GINN-KAN on datasets constructed using these equations. We create these datasets by sampling 2000 points randomly for each input variable and generating the outputs using the ground truth equation.

\textbf{Symbolic regression benchmark.}
We perform the evaluation of GINN-KAN on the popular Feynman symbolic regression benchmark datasets using SRBench, a popular benchmark for symbolic regression methods \cite{la_cava_contemporary_2021}. This dataset contains 114 datasets with known ground truth equations and is often used for evaluating symbolic regression methods \cite{udrescu_ai_2020}.

\textbf{Implementation details}. The performance of GINN-KAN were compared with GINN, KAN and other contemporary symbolic regression methods on the Feynman symbolic regression benchmark dataset. The architecture of each model is as follows:

\begin{itemize}
    \item GINN: It consists of parallel PTAs, starting with $1$ block and growing up to $4$ during training.  These parallel PTA blocks are followed by a single linear output neuron that combines their outputs.
    \item KAN: We use a $2$-layer KAN with $5$ nodes in the hidden layer. With this architecture, there are $15$ B-Spline activation functions that need to be estimated ($10$ functions from the input layer to the hidden layer and $5$ functions from the hidden layer to the output layer). Two key parameters of the activation function, the number of intervals $G$ and the order $k$ of the piecewise polynomial, are chosen as $5$ and $3$ respectively, which aligns with the default choices of the KAN's authors. \cite{liu_kan_2024}.
    \item GINN-KAN: The model consists of two parallel GINN blocks, followed by a 2-layer KAN with 5 nodes in its hidden layer. It's important to note that the structures of the GINN and KAN components used in this GINN-KAN model are identical to their standalone counterparts. This makes the comparison of the performance of GINN-KAN to that of a standalone KAN a fair evaluation, as the number of parameters in GINN-KAN is only marginally larger than in KAN alone. This is due to the GINN component using very few parameters, making the increase in total parameter count negligible.
    \item Existing symbolic regression methods: The parameters and architecture of each symbolic regression method follow the choices used in SR Bench \cite{la_cava_contemporary_2021}.
\end{itemize}
\textbf{Evaluation metrics}. After training the models on each SR dataset, we measure the performance using the Mean Squared Error (MSE) and the coefficient of determination $R^2$ evaluated on the test set. The $R^2$ is aggregated by calculating the percentage of datasets with $R^2 > 0.99$. This metric is also used in SRBench, a widely used symbolic regression benchmark \cite{la_cava_contemporary_2021}. We calculate the MSE rank across GINN, KAN and GINN-KAN and average it to calculate the mean rank on the LP / Non-LP subsets of datasets, allowing us to compare performance within this subgroup of models.


\subsection{GINN-KAN Augmented PINNs}
\textbf{Partial differential equations}. The performance of GINN-KAN is evaluated on 15 PDEs, which are shown in the Appendix. The first 7 equations, include both linear and nonlinear PDEs commonly used to model various physical phenomena. These equations vary in complexity and involve different operations (e.g., multiplication, division, addition) and basis functions (e.g., $\sin$, $\exp$, $x^2$). The last 5 equations are designed with LP ground truth analytical equations.

\textbf{Implementation details}. The performance of GINN-KAN was compared with GINN, KAN, and a fully connected (FC) network. The FC network consists of $5$ layers, each containing $32$ neurons. The hyperbolic tangent function is used as the activation mechanism throughout the network. For GINN, KAN, and GINN-KAN, the architecture is the same as described above.

When training each model, we randomly sample 2500 collocation points $(x,t)$ in the spatial-temporal domain as the input. All models are trained to minimize the loss function in Eq. \ref{eq:physics_loss} using the Adam optimizer with a learning rate of 0.01. 
It is worth highlighting that one characteristic of PINNs is their ability to perform unsupervised learning, eliminating the need for labelled data \cite{raissi_physics-informed_2018}. Similarly, in this paper, all models operate in an unsupervised manner, relying solely on 2500 sampled points without prior knowledge of their output values.

\textbf{Evaluation metrics}. After training the models, the performance is evaluated by measuring the MSE between the prediction solution and the analytical solution among all randomly sampled 2500 points (sampled during the training stage). For all equations, each model will be run five times with different preset random seeds. The median MSE and mean rank across all equations of each model are then calculated. These aggregate metrics provide an overall performance assessment for each model across the various PDEs tested.

All experiments were performed on a computing cluster equipped with a 32-core, 2.90GHz Intel Xeon Gold 6326 CPU and a single NVIDIA A-100 GPU for each experiment.

\section{Results and Discussion}
\subsection{GINN-KAN}

\begin{table}[ht]

\begin{tabular}{|l|c|c|c|c|}
\hline
\textbf{Equation}                  & \textbf{GINN} & \textbf{KAN} & \textbf{GINN-KAN} \\ \hline \hline
$\sin(x_1) + x_2$                      & 1.07E-01          & 3.90E-04      & \textbf{1.70E-04}  \\ \hline
$\sin(x_1) + \sin(x_2)$                  & 1.97E-01          & 5.33E-04      & \textbf{5.66E-05}  \\ \hline
$x_1*x_2^2$           & \textbf{3.55E-09} & 1.08E-02      & 5.70E-04           \\ \hline
$2x_1 + 3x_2^2 + x_1x_2$            & \textbf{1.40E-03} & 4.05E-03      & 1.41E-02           \\ \hline
$\sin(x_1*x_2)$                        & 4.50E-01          & 1.21E-01      & \textbf{1.30E-04}  \\ \hline
$\sin(x_1*x_2) + x_1$                    & 4.48E-01          & 1.68E-01      & \textbf{9.38E-02}  \\ \hline
$x_1*x_2^2 + \sin(x_1)$ & 8.06E-02          & 1.18E-02      & \textbf{7.76E-03}  \\ \hline
$\ln(x_1*x_2 + x_1)$                     & 1.48E-05          & 2.90E-04      & \textbf{5.46E-07}  \\ \hline
\end{tabular}

\caption{Performance comparison interpretable networks in terms of MSE on datasets generated by LP and Non-LP equations with and without multiplicative terms. This showcases the limitations of GINN in learning non-LP equations and the limitation of KAN in learning equations with multiplicative terms.}
\label{tab:toy_examp}

\end{table}

\begin{figure*}[ht]
    \centering
    \includegraphics[width=\textwidth]{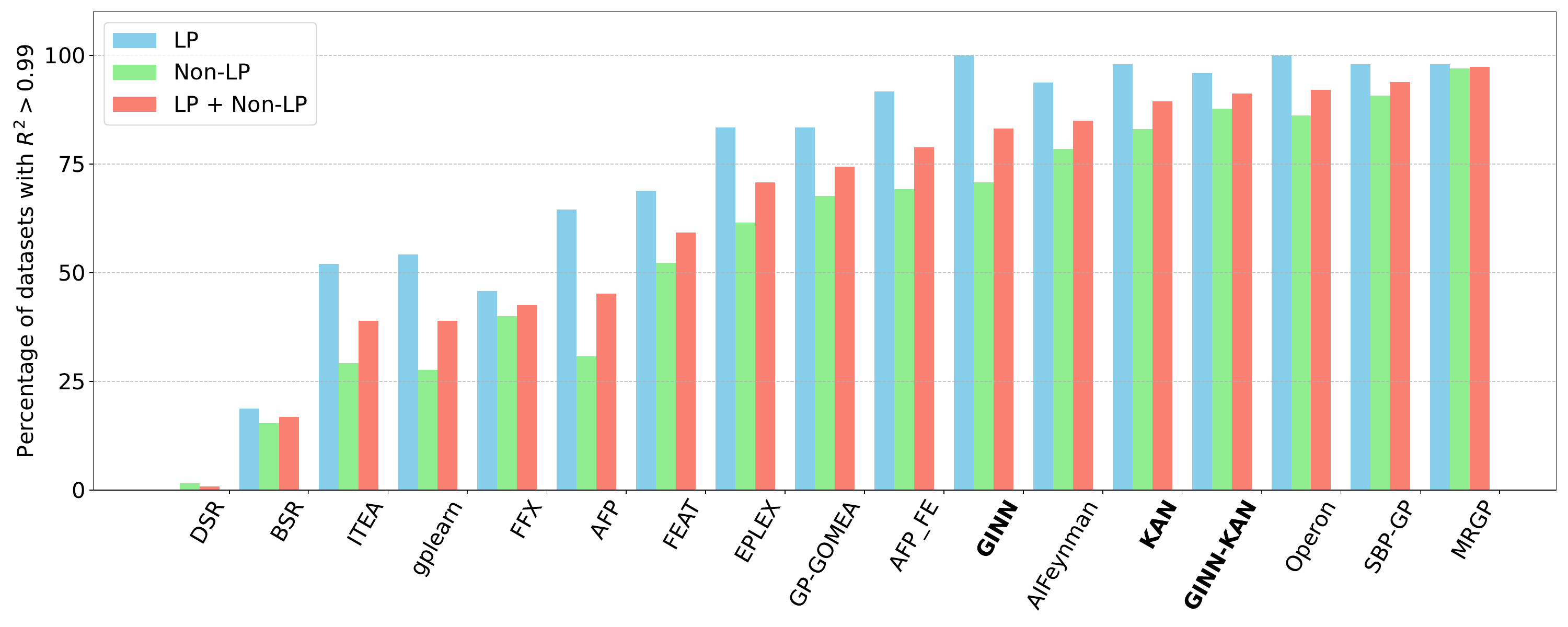}
    \caption{Performance of GINN, KAN and GINN-KAN when compared to existing symbolic regression methods. GINN, KAN and GINN-KAN are the only methods that can be trained using backpropagation. Experiments are performed on 114 SR datasets from the Feynman symbolic regression benchmark. The percentage of datasets with $R^2 > 0.99$ is reported. The bold labels show the methods that can be trained using backpropagation.}
    \label{fig: feynman_sr}
\end{figure*}

We show the results on several synthetic datasets generated in Table \ref{tab:toy_examp}. The first two equations are non-LP equations with no multiplicative terms. On these datasets, GINN performs poorly but KANs and GINN-KAN both perform well on these datasets. The next two equations are LP equations, on which GINN performs well, but KANs do not perform well. The last 4 equations are non-LP equations with multiplicative terms. On these datasets, GINN-KAN performs more than an order of magnitude better than the next best method. These results further confirm the limitations of KANs when approximating multiplicative terms within variables. Although theoretically possible, in practice KANs seem to perform worse when the underlying function contains a multiplication. Moreover, as expected due to the specialized nature of its architecture, GINN does not perform well when the underlying ground truth equation does not follow a LP form.

Fig. \ref{fig: feynman_sr} presents the results of GINN-KAN, comparing its performance against GINN, KAN, and other symbolic regression methods on the Feynman symbolic regression benchmark dataset. Here we show the $R^2 > 0.99$ accuracy, highlighting that GINN-KAN outperforms the majority of the other symbolic regression methods against which it was compared. Moreover, GINN-KAN performs better than both GINN and KAN, which indicates that the ensemble performs better than each of the individual models. 

Since the architecture of GINN allows it to better approximate datasets with LP ground truths, we further analyze these results in the subgroups of datasets which have LP and Non-LP ground truth equations. We compare the mean ranks of GINN, KAN and GINN-KAN in terms of the MSE for the two subsets and show them in Table \ref{mean_rank_feynman}. GINN performs the best on the LP subset of data, and relatively worse on the Non-LP subset. GINN-KAN clearly outperforms both the other methods when the ground truth equation is a Non-LP and shows the best overall performance across all types of equations. This further indicates that the ensemble GINN-KAN combines the strengths of these two methods, performing better than each of the individual models. 
    
Although a few methods outperform GINN-KAN in terms of the $R_2 > 0.99$ accuracy, we note that none of these methods can be trained using backpropagation, and therefore have limited applicability in developing more powerful interpretable machine learning methods through pipelining. Moreover, such methods cannot be seamlessly integrated with existing machine learning pipelines.

\begin{table}[ht]
\centering
\begin{tabular}{ l l c c }
\hline\hline
\textbf{Method}   & \textbf{LP}     & \textbf{Non-LP} & \textbf{LP + Non-LP} \\  \hline\hline
GINN     & \textbf{1.1875} & 2.0562 & 1.6814 \\ \hline
KAN      & 2.8958 & 2.6154 & 2.7345 \\ \hline
GINN-KAN & 1.9167 & \textbf{1.3385} & \textbf{1.5840} \\ \hline
\hline
\end{tabular}
\caption{GINN, KAN and GINN-KAN on 114 SR datasets from the Feynman symbolic regression benchmark. We show the mean rank of each method when trained on the subset datasets with LP/Non-LP ground truth equations.}  
\label{mean_rank_feynman}
\end{table}

\subsection{GINN-KAN Augmented PINNs}

\begin{table*}[h!]
\centering
\begin{tabular}{ l c c c c c }
\hline\hline
\textbf{PDE} & \textbf{GINN-KAN} & \textbf{GINN} & \textbf{KAN} & \textbf{FC Network} & \textbf{Operators} \\
\hline\hline
Inviscid Burgers' & \textcolor{blue}{3.07E-03} & 5.59E-03 & \textcolor{blue}{\textbf{3.70E-04}} & 9.35E-03 &  $ + / * / x / (1/x)$ \\
\hline
Convection 1 & \textcolor{blue}{5.49E-01} & \textcolor{blue}{\textbf{5.05E-01}} & 9.03E-01 & 9.11E-01 & $+ / \sin(x)$\\
\hline
Convection 2 & 6.75E-04 & 5.15E-01 & \textcolor{blue}{\textbf{5.59E-04}} & \textcolor{blue}{3.74E-03} & $+ / \sin(x)$\\
\hline
Diffusion & \textcolor{blue}{2.70E-01} & \textcolor{blue}{\textbf{1.09E-01}} & 3.56E+00 & 5.23E-01 & $ * / e^x / \sin(x)$ \\
\hline
Fokker-Planck & \textcolor{blue}{8.81E-01} & \textcolor{blue}{\textbf{4.15E-01}} & 9.98E-01 & 9.75E-01 & $+ / * / e^x$ \\
\hline
Reaction & 4.28E-02 & 1.98E-01 & \textcolor{blue}{\textbf{4.02E-02}} & \textcolor{blue}{4.17E-02} & $+ / * / e^x / (1/x)$\\
\hline
Telegraph & \textcolor{blue}{2.84E-03} & 5.14E-02 & \textcolor{blue}{\textbf{3.35E-04}} & 9.38E-01 & $ + / e^x $ \\
\hline
Wave & \textcolor{blue}{\textbf{7.20E-02}} & \textcolor{blue}{9.02E-02} & 2.01E-01 & 1.54E-01 & $ * / \sin(x) $\\
\hline
Toy 1 & \textcolor{blue}{\textbf{1.11E-03}} & 1.09E-01 & \textcolor{blue}{3.51E-03} & 1.05E-02 & $ +/ */ \cos(x) $\\
\hline
Toy 2 & \textcolor{blue}{1.54E-05} & 1.31E-04 & \textcolor{blue}{\textbf{6.07E-07}} & 9.90E-05 & $ + / e^x / (1/x) $\\
\hline
LP 1 & 3.15E-02 & \textcolor{blue}{\textbf{2.06E-07}} & \textcolor{blue}{1.07E-03} & 5.77E-01 & $+ / x / x^2$\\
\hline
LP 2 & 9.61E-02 & 9.92E-02 & \textcolor{blue}{9.10E-02} & \textcolor{blue}{\textbf{7.59E-02}} & $+ / * $\\
\hline
LP 3 & 1.29E-01 & \textcolor{blue}{2.57E-02} & \textcolor{blue}{\textbf{6.90E-05}} & 1.75E+02 & $+ / x^2$\\
\hline
LP 4 & \textcolor{blue}{7.65E-03} & 1.10E-02 & \textcolor{blue}{\textbf{6.15E-05}} & 1.49E+01 & $+ / x^2$\\
\hline
LP 5 & 2.83E-02 & \textcolor{blue}{\textbf{2.14E-04}} & \textcolor{blue}{2.56E-03} & 7.66E-02 & $+ / x^2 / x^3 / x^4$\\
\hline
\textbf{Average (All)} & \textbf{1.41E-01} & 1.42E-01 & 3.87E-01 & 1.29E+01 \\
\hline
\textbf{Average (Excluding LP 1-5)} & \textbf{1.82E-01} & 2.00E-01 & 5.71E-01 & 3.57E-01 \\
\hline
\textbf{Mean Rank (All)} & 2.20 & 2.53 & \textbf{2.00} & 3.27 \\
\hline
\textbf{Mean Rank (Excluding LP 1-5)} & \textbf{1.90} & 2.70 & 2.20 & 3.20 \\
\hline\hline
\end{tabular}
\caption{Comparison of PINN performance with different types of interpretable neural networks. The compared equations are available in the supplementary material\ref{appendix}. The operators show the unique set of operators that are present in each equation. We show the median MSE across 5 experiments with different random seeds. Results indicate that equations with multiplications are more difficult for KANs, and equations which do not take an LP form are more difficult for GINN. The best performing method is shown in bold blue colour while the second best method is shown in regular blue.}
\label{pinn_result}
\end{table*}

The results of PINN performance, measured in MSE, with different types of interpretable neural networks, are presented in Table \ref{pinn_result}. Both KAN, GINN and GINN-KAN outperform the FC network in most PDEs. GINN performs reasonably well in solving LPs, as evidenced by the LP equations $1-5$. However, its performance varies when dealing with non-LP PDEs. For instance, its performance is much worse than the GINN-KAN and KAN when solving Convection $2$, Reaction, and Toy $1$ and $2$ equations,  by orders of magnitude.

KAN performs best in $7$ out of $15$ equations, demonstrating its capability in solving PDEs. However, compared to GINN-KAN, it performs worse in recovering analytical solutions involving multiplication (such as the Wave, Diffusion and Toy $1$ equations) due to the Kolmogorov-Arnold representation theorem, which treats functions as the summation of multiple univariate functions. This approach makes it challenging to learn multiplication (as discussed earlier). 

In contrast, GINN-KAN combines the strengths of GINNs and KANs, resulting in the best overall performance across all PDEs. GINN-KAN shows the most consistent performance, with the lowest average MSE ($1.41$E$-01$) and the second-best mean rank ($2.20$) for all equations and best mean rank when excluding the LP equations. It performs well in complex PDEs like Convection $1$ and Wave equations, while maintaining competitive performance in LP problems. This demonstrates GINN-KAN's effectiveness in handling a wide range of PDEs.

Notably, GINN-KAN successfully addresses KAN's limitations with multiplicative terms, as evidenced by its performance in equations like Wave and Diffusion. Unlike GINN, which shows proficiency in LP problems but struggles with some non-LP equations, GINN-KAN maintains good performance across both types of problems. This balanced and robust performance across various PDE types further supports the claim that GINN-KAN is a powerful method for solving a wide range of PDEs.

\section{Conclusion}

In this work, we first evaluate interpretable neural networks, specifically focusing on the recently introduced GINN and KANs. While both methods have demonstrated interpretability, they exhibit limitations in learning certain types of functions. Our evaluation shows that GINNs do not perform well on datasets governed by non-LP equations, while KANs do not perform well on datasets governed by equations with multiplications. To address these limitations, we propose a novel interpretable neural network, GINN-KAN, which combines the strengths of both GINN and KANs. Experiments conducted with this novel network on the Feynman symbolic regression benchmark datasets show that GINN-KAN outperforms both GINN and KANs on datasets with known ground truth equations. We then apply GINN-KANs to physics-informed neural networks, showing that they can add interpretability to PINNs. By performing experiments on 15 differential equations, we demonstrate that this interpretable PINN not only adds interpretability but also improves the performance in solving differential equations when compared with traditional PINNs.

Interpretability is a crucial aspect often overlooked in favor of performance in neural networks. However, methods like GINN-KAN have the potential to be as effective as black-box MLPs while also being interpretable. Since GINN-KAN can be trained using backpropagation, it can be seamlessly integrated into existing machine learning pipelines with minimal adjustments to the training strategy.

Despite its advantages, GINN-KAN shares the limitation of GINN in being restricted to inputs with positive values. This can be mitigated by shifting all inputs to the positive range. Future research should explore more robust architectures that maintain interpretability while overcoming this limitation. Additionally, better regularization techniques are needed to enable GINN-KAN to accurately discover concise ground truth equations that describe the data.

In conclusion, GINN-KAN represents a significant step forward in developing interpretable neural networks, offering a promising balance between interpretability and performance. Its ability to integrate into existing machine learning frameworks and its potential applications in physics-informed neural networks highlight its importance in advancing the field.

\bibliography{references}

\begin{thebibliography}{16}
\providecommand{\natexlab}[1]{#1}

\bibitem[{Ali et~al.(2023)Ali, Abuhmed, El-Sappagh, Muhammad, Alonso-Moral, Confalonieri, Guidotti, Del~Ser, Díaz-Rodríguez, and Herrera}]{ali_explainable_2023}
Ali, S.; Abuhmed, T.; El-Sappagh, S.; Muhammad, K.; Alonso-Moral, J.~M.; Confalonieri, R.; Guidotti, R.; Del~Ser, J.; Díaz-Rodríguez, N.; and Herrera, F. 2023.
\newblock Explainable {Artificial} {Intelligence} ({XAI}): {What} we know and what is left to attain {Trustworthy} {Artificial} {Intelligence}.
\newblock \emph{Information Fusion}, 99: 101805.

\bibitem[{Bozorgasl and Chen(2024)}]{bozorgasl_wav-kan_2024}
Bozorgasl, Z.; and Chen, H. 2024.
\newblock Wav-{KAN}: {Wavelet} {Kolmogorov}-{Arnold} {Networks}.
\newblock ArXiv:2405.12832 [cs, eess, stat].

\bibitem[{La~Cava et~al.(2021)La~Cava, Orzechowski, Burlacu, de~França, Virgolin, Jin, Kommenda, and Moore}]{la_cava_contemporary_2021}
La~Cava, W.; Orzechowski, P.; Burlacu, B.; de~França, F.~O.; Virgolin, M.; Jin, Y.; Kommenda, M.; and Moore, J.~H. 2021.
\newblock Contemporary {Symbolic} {Regression} {Methods} and their {Relative} {Performance}.
\newblock In \emph{Thirty-fifth {Conference} on {Neural} {Information} {Processing} {Systems} {Datasets} and {Benchmarks} {Track} ({Round} 1)}.
\newblock ArXiv: 2107.14351.

\bibitem[{Liu et~al.(2024)Liu, Wang, Vaidya, Ruehle, Halverson, Soljačić, Hou, and Tegmark}]{liu_kan_2024}
Liu, Z.; Wang, Y.; Vaidya, S.; Ruehle, F.; Halverson, J.; Soljačić, M.; Hou, T.~Y.; and Tegmark, M. 2024.
\newblock {KAN}: {Kolmogorov}-{Arnold} {Networks}.
\newblock ArXiv:2404.19756 [cond-mat, stat].

\bibitem[{Lundberg and Lee(2017)}]{lundberg_unified_2017}
Lundberg, S.~M.; and Lee, S.-I. 2017.
\newblock A unified approach to interpreting model predictions.
\newblock In \emph{Advances in neural information processing systems}.

\bibitem[{Podina, Eastman, and Kohandel(2023)}]{podina_universal_2023}
Podina, L.; Eastman, B.; and Kohandel, M. 2023.
\newblock Universal {Physics}-{Informed} {Neural} {Networks}: {Symbolic} {Differential} {Operator} {Discovery} with {Sparse} {Data}.
\newblock In \emph{Proceedings of the 40th {International} {Conference} on {Machine} {Learning}}, 27948--27956. PMLR.
\newblock ISSN: 2640-3498.

\bibitem[{Raissi, Perdikaris, and Karniadakis(2017)}]{raissi_physics_2017}
Raissi, M.; Perdikaris, P.; and Karniadakis, G.~E. 2017.
\newblock Physics {Informed} {Deep} {Learning} ({Part} {I}): {Data}-driven {Solutions} of {Nonlinear} {Partial} {Differential} {Equations}.
\newblock ArXiv: 1711.10561.

\bibitem[{Raissi, Perdikaris, and Karniadakis(2018)}]{raissi_physics-informed_2018}
Raissi, M.; Perdikaris, P.; and Karniadakis, G.~E. 2018.
\newblock Physics-{Informed} {Neural} {Networks}: {A} {Deep} {Learning} {Framework} for {Solving} {Forward} and {Inverse} {Problems} {Involving} {Nonlinear} {Partial} {Differential} {Equations}.
\newblock Technical report.

\bibitem[{Ranasinghe et~al.(2024)Ranasinghe, Senanayake, Seneviratne, Premaratne, and Halgamuge}]{ranasinghe_ginn-lp_2024}
Ranasinghe, N.; Senanayake, D.; Seneviratne, S.; Premaratne, M.; and Halgamuge, S. 2024.
\newblock {GINN}-{LP}: {A} {Growing} {Interpretable} {Neural} {Network} for {Discovering} {Multivariate} {Laurent} {Polynomial} {Equations}.
\newblock \emph{Proceedings of the AAAI Conference on Artificial Intelligence}, 38(13): 14776--14784.
\newblock Number: 13.

\bibitem[{Ribeiro, Singh, and Guestrin(2016)}]{ribeiro_why_2016}
Ribeiro, M.~T.; Singh, S.; and Guestrin, C. 2016.
\newblock "{Why} {Should} {I} {Trust} {You}?": {Explaining} the {Predictions} of {Any} {Classifier}.
\newblock In \emph{Proceedings of the 22nd {ACM} {SIGKDD} {International} {Conference} on {Knowledge} {Discovery} and {Data} {Mining}}, 1135--1144. New York, NY, USA: ACM.
\newblock ISBN 978-1-4503-4232-2.

\bibitem[{Sahoo, Lampert, and Martius(2018)}]{sahoo_learning_2018}
Sahoo, S.~S.; Lampert, C.~H.; and Martius, G. 2018.
\newblock Learning {Equations} for {Extrapolation} and {Control}.
\newblock ArXiv: 1806.07259.

\bibitem[{Selvaraju et~al.(2016)Selvaraju, Cogswell, Das, Vedantam, Parikh, and Batra}]{selvaraju_grad-cam_2016}
Selvaraju, R.~R.; Cogswell, M.; Das, A.; Vedantam, R.; Parikh, D.; and Batra, D. 2016.
\newblock Grad-{CAM}: {Visual} {Explanations} from {Deep} {Networks} via {Gradient}-based {Localization}.
\newblock ArXiv: 1610.02391.

\bibitem[{Udrescu and Tegmark(2020)}]{udrescu_ai_2020}
Udrescu, S.-M.; and Tegmark, M. 2020.
\newblock {AI} {Feynman}: {A} physics-inspired method for symbolic regression.
\newblock Technical report.

\bibitem[{Vaca-Rubio et~al.(2024)Vaca-Rubio, Blanco, Pereira, and Caus}]{vaca-rubio_kolmogorov-arnold_2024}
Vaca-Rubio, C.~J.; Blanco, L.; Pereira, R.; and Caus, M. 2024.
\newblock Kolmogorov-{Arnold} {Networks} ({KANs}) for {Time} {Series} {Analysis}.
\newblock ArXiv:2405.08790 [cs, eess].

\bibitem[{Xu et~al.(2024)Xu, Chen, Li, Yang, Wang, Hu, and Ngai}]{xu_fourierkan-gcf_2024}
Xu, J.; Chen, Z.; Li, J.; Yang, S.; Wang, W.; Hu, X.; and Ngai, E. C.-H. 2024.
\newblock {FourierKAN}-{GCF}: {Fourier} {Kolmogorov}-{Arnold} {Network} -- {An} {Effective} and {Efficient} {Feature} {Transformation} for {Graph} {Collaborative} {Filtering}.
\newblock ArXiv:2406.01034 [cs].

\bibitem[{Zhou et~al.(2016)Zhou, Khosla, Lapedriza, Oliva, and Torralba}]{zhou_learning_2016}
Zhou, B.; Khosla, A.; Lapedriza, A.; Oliva, A.; and Torralba, A. 2016.
\newblock Learning {Deep} {Features} for {Discriminative} {Localization}.
\newblock In \emph{2016 {IEEE} {Conference} on {Computer} {Vision} and {Pattern} {Recognition} ({CVPR})}, 2921--2929. Las Vegas, NV, USA: IEEE.
\newblock ISBN 978-1-4673-8851-1.

\end{thebibliography}
\appendix

\newpage
\section{Appendix}
\label{appendix}

\renewcommand{\arraystretch}{1.5}
\begin{sidewaystable}[h]
\hspace{-2cm}
\vspace{8.5cm}
\begin{tabular}{|c|c|c|c|c|}
\hline
\textbf{ } & \textbf{PDEs} & \textbf{Initial Conditions} & \textbf{Boundary Conditions} & \textbf{True Solution} \\ \hline

Inviscid Burgers'      & \makecell{$\frac{\partial u}{\partial t}+u \frac{\partial u}{\partial x}=0$ \\ $x \in [0,2]$, $t \in [0,1]$}  & $u(x,0)=2x+1$ & $u(0,t)=0$ & $u(x,t)=\frac{2x+1}{2t+1}$\\ \hline

Convection 1  & \makecell{$\frac{\partial u}{\partial t}+ 30\frac{\partial u}{\partial x}=0$ \\$x \in [0,2\pi]$, $t \in [0,1]$} & $u(x,0)=\sin(x)$& $u(0,t)=u(2\pi,t)$& $u(x,t)=\sin(x-30 t)$   \\ \hline

Convection  2 & \makecell{$\frac{\partial u}{\partial t}+5\frac{\partial u}{\partial x}=0$ \\$x \in [0,2\pi]$, $t \in [0,1]$} & $u(x,0)=\sin(x)$& $u(0,t)=u(2\pi,t)$& $u(x,t)=\sin(x-5t)$   \\ \hline

Diffusion    & \makecell{$\frac{\partial u}{\partial t}-\frac{\partial^2 u}{\partial x^2}+e^{-t}\sin(\pi x)(1-\pi)^2=0$ \\ $x \in [-1,1]$, $t \in [0,1]$}  & $u(x,0)=\sin(\pi x)$& $u(-1,t)=u(1,t)=0$ & $u(x,t)=e^{-t}\sin(\pi x)$   \\ \hline

Fokker-Planck & \makecell{$\frac{\partial u}{\partial t}- (x+1)\frac{\partial u}{\partial x} -x^2e^t\frac{\partial^2 u}{\partial x^2}=0$ \\ $x \in [0,1]$, $t \in [0,1]$} &$u(x,0)=x+1$& $u(0,t)=e^{t}$& $u(x,t)=(x+1)e^{t}$   \\ \hline

Reaction     & \makecell{$\frac{\partial u}{\partial t}+3u(1-u)=0$; $h(x)=e^{-\frac{(x-\pi)^2}{2(\pi/4)^2}}$ \\$x \in [0,2\pi]$, $t \in [0,1]$} & $u(x,0)=h(x)$ & $u(0,t)=u(2\pi,t)$ &  $u(x,t)=\frac{h(x)e^{3t}}{h(x)e^{3t}+1-h(x)}$  \\ \hline

Telegraph    & \makecell{$\frac{\partial^2 u}{\partial t^2}+2\frac{\partial u}{\partial t}+u  -\frac{\partial^2 u}{\partial x^2}=0$ \\ $x \in [0,1]$, $t \in [0,1]$} & $\frac{\partial u}{\partial t}(x,0)=-1$; $u(x,0)=e^{x}+1$ & $u(0,t)=e^{-t}+1$ & $u(x,t)=e^{x}+e^{-t}$   \\ \hline

Wave         & \makecell{$\frac{\partial^2 u}{\partial t^2}-\frac{\partial^2 u}{\partial x^2}=0$ \\ $x \in [0,\pi]$, $t \in [0,3]$} & $\frac{\partial u}{\partial t}(x,0)=\sin(x)$; $u(x,0)=0$& $u(0,t)=u(\pi,t)=0$ & $u(x,t)=\sin(x)\sin(t)$   \\ \hline

Toy 1         & \makecell{$\frac{\partial u}{\partial t}-\frac{x+2}{t+1}\frac{\partial u}{\partial x}=0$ \\ $x \in [-1,1]$, $t \in [0,1]$} & $u(x,0)=\cos(x+2)$& $u(1,t)=\cos(3(t+1))$ & $u(x,t)=\cos((t+1)(x+2))$   \\ \hline

Toy 2        & \makecell{$\frac{\partial u}{\partial t}+\frac{x}{t+1}\frac{\partial u}{\partial x}=0$ \\ $x \in [0,2]$, $t \in [0,1]$} & $u(x,0)=e^{-x}$& $u(0,t)=1$ & $u(x,t)=e^{-\frac{x}{t+1}}$   \\ \hline

LP 1         & \makecell{$\frac{\partial u}{\partial t}-4t\frac{\partial u}{\partial x}=0$  \\ $x \in [0,2]$, $t \in [0,1]$} & $u(x,0)=3x+1$ & $u(0,t)=6t^2+1$ & $u(x,t)=1+3x+6t^2$\\ \hline

LP 2         & \makecell{ $\frac{\partial u}{\partial t}-\frac{x}{1+t}\frac{\partial u}{\partial x}=0$ \\ $x \in [0,2]$, $t \in [0,1]$}  & $u(x,0)=x+1$ & $u(0,t)=1$; $u(1,t)=2+t$ & $u(x,t)=1+x(1+t)$    \\ \hline

LP 3         & \makecell{ $x\frac{\partial u}{\partial t}-25\frac{\partial u}{\partial x}=0$ \\ $x \in [0,2]$, $t \in [0,1]$}  & $u(x,0)=x^2+1$ & $u(0,t)=50t+1$ & $u(x,t)=1+x^2+50t$    \\ \hline

LP 4         &  \makecell{$6x\frac{\partial u}{\partial t}-\frac{\partial u}{\partial x}=0$ \\ $x \in [0,2]$, $t \in [0,1]$}  & $u(x,0)=15x^2+1$ & $u(0,t)=5t+1$ & $u(x,t)=1+15x^2+5t$    \\ \hline

LP 5         & \makecell{$\frac{\partial u}{\partial t}-\frac{1}{2}\frac{\partial^2 u}{\partial x^2}+3x-4t^3=0$ \\ $x \in [0,2]$, $t \in [0,1]$}  & $u(x,0)=t^4+t+1$ & $u(0,t)=x^3+x^2+1$ & $u(x,t)=1+x^2+x^3+t+t^4$    \\ \hline
\end{tabular}

\caption{PDE information for experiments}
\end{sidewaystable}

\end{document}